\documentclass[10pt,twocolumn,letterpaper]{article}

\usepackage{cvpr}              

\usepackage{graphicx}
\usepackage{booktabs}

\usepackage[accsupp]{axessibility}  

\usepackage[pagebackref,breaklinks,colorlinks,allcolors=cvprblue]{hyperref}

\usepackage{orcidlink}
\usepackage[accsupp]{axessibility}  

\definecolor{ao(english)}{rgb}{0.0, 0.5, 0.0}

\usepackage{array}
\newcolumntype{P}[1]{>{\centering\arraybackslash}p{#1}}
\newcolumntype{L}[1]{>{\raggedright\arraybackslash}p{#1}}
\newcolumntype{C}[1]{>{\centering\arraybackslash}p{#1}}

\definecolor{cvprblue}{rgb}{0.21,0.49,0.74}
\def\paperID{52} 
\def\confName{CVPR}
\def\confYear{2026}

\title{HierSum: A Global and Local Attention Mechanism for Video Summarization}
\author{Apoorva Beedu$^\dagger$$^\mathsection$\thanks{Work done while a student at Georgia Institute of Technolgy.} , Irfan Essa$^{\dagger}$\\
$^\dagger$Georgia Institute of Technology, $^\mathsection$Rivian and Volkswagen Group\\
{\tt\small abeedu3, irfan@gatech.edu}
}

\begin{document}
\maketitle
\begin{abstract}
Video summarization creates an abridged version (i.e., a summary) that provides a quick overview of the video while retaining pertinent information. 
In this work, we focus on summarizing instructional videos and propose a method for breaking down a video into meaningful segments, each corresponding to essential steps in the video. 
We propose \textbf{HierSum}, a hierarchical approach that integrates fine-grained local cues from subtitles with global contextual information provided by video-level instructions. 
Our approach utilizes the ``most replayed" statistic as a supervisory signal to identify critical segments, thereby improving the effectiveness of the summary. 
We evaluate on benchmark datasets such as TVSum, BLiSS, Mr.HiSum, and the WikiHow test set, and show that HierSum consistently outperforms existing methods in key metrics such as F1-score and rank correlation. 
We also curate a new multi-modal dataset using WikiHow and EHow videos, and associated articles containing step-by-step instructions. 
Through extensive ablation studies, we demonstrate that training on this dataset significantly enhances summarization on the target datasets. 
The code and dataset will be available at \href{https://github.com/ApoorvaBeedu/HierSum}{https://github.com/ApoorvaBeedu/HierSum}.
\end{abstract}    
\section{Introduction}
\label{sec:intro}

Finding the right video content online continues to be a challenge. 
Simple search queries such as ``How to make an omelet?'' give hundreds of results, with videos ranging from a couple of minutes to more than ten minutes. 
This makes it harder to identify relevant videos from the hundreds of suggestions. 
Extracting and identifying essential and relevant video segments for a given query significantly, i.e., Video Summarization, improves the experience of watching task-related videos.
It seeks to automatically select the most critical segments in a video, condensing the content while preserving essential information.

Uni-modal techniques such as keyframe extraction~\cite{baghel2020image,tan2024large}, scene segmentation~\cite{demir2015video,otani2017video}, and keyframe importance voting~\cite{song2015tvsum} have been explored for summarization. 
Recently, there has been increasing interest in multimodal summarization~\cite{he2023align,yan2023unloc,lin2023univtg}, which identifies keyframes by integrating information across multiple modalities~\cite{lin2023univtg,narasimhan2021clip,han2023shot2story20k} or by leveraging textual data for alignment~\cite{narasimhan2022tl,he2023align}, typically within a single-stage framework. Most of these approaches are trained and evaluated on datasets like TVSum~\cite{song2015tvsum} and SumMe~\cite{gygli2014creating}, which offer frame-level importance annotations. However, due to the high cost of manual annotation, these datasets are relatively small and cover only a narrow spectrum of events or actions.
\begin{figure}
    \centering
    \includegraphics[width=1.0\columnwidth]{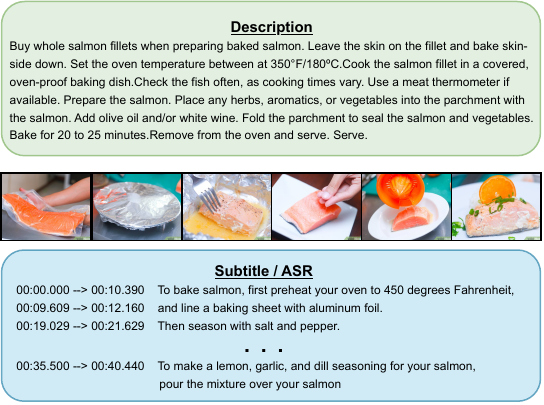}
    \caption{
    Instructional video on ``How to Bake Salmon" from WikiHow showing Automatic Speech Recognition (ASR) text and instructions from the associated WikiHow article. 
    Subtitles sometimes miss important instructions such as ``Leave the skin on", whereas global descriptions lack video-text alignment.}
    \label{fig:template}
\end{figure}

To address the issue of scale and diversity of tasks in standard datasets, recent methods \textit{generate} pseudo-summaries. 
He~\etal\cite{he2023align} extract keyframes by matching frames with video thumbnails.
TL;DW~\cite{narasimhan2022tl} generates pseudo summaries from a large corpus of instructional videos by leveraging consistency of essential steps across different videos of the same task. 
However, this approach tends to capture only the most frequent steps, overlooking unique yet important actions specific to individual videos \cite{narasimhan2022tl}.
In contrast, \cite{sul2024mr} utilizes replay statistics from YouTube videos to generate scores, with these statistics highlighting key moments; He~\etal\cite{sul2024mr} further demonstrate that these measures effectively aid video summarization.
Although these methods partially mitigate the challenges posed by small-scale datasets, they fail to exploit the \textit{detailed instructions that instructional videos provide}.

To tackle this issue, we develop a dataset with \textit{both clip-level subtitles and video-level instructions}, and use it during training. 
The essential events in a video are often captured by either Automatic Speech Recognition (ASR) generated subtitles, video-level descriptions, or combination of both. 
As illustrated in \autoref{fig:template}, the description contains all the relevant information about the task ``\textit{How to Bake Salmon}" including additional details such as \textit{Leave the skin on}, while the instructions provided by the subtitles are more fine-grained.
We demonstrate that training a hierarchical model across both global and local levels, augmented with \textsl{most replayed} scores, effectively captures all key segments in a video, resulting in improved summarization performance.
This approach results in a nearly 2\% improvement on the rank coefficient metric for TVSum, 2\% improvement on F1-score for Mr.HiSum, and 1\% improvement on cosine similarity for the BLiSS dataset.

In summary, our contributions are: 
\begin{itemize}
    \item We introduce HierSum, a novel hierarchical framework that seamlessly integrates local features from subtitles with global video-level instructions. 
    By employing a parent-child training strategy, our method effectively captures both detailed subtitle-level information and long-range contextual cues.
    \item We introduce a new dataset of instructional videos annotated with multi-granularity information.
    This enriches the training process by providing diverse instructional signals, rather than simply expanding the size of the pre-training data.
    \item Our experiments demonstrate that incorporating highlight labels -- derived from \textsl{most replayed} statistics, in conjunction with text inputs significantly enhances video summarization performance. 
Additionally, ablation studies and extensive experiments indicate that pre-training on the most replayed statistics datasets yields an improvement of approximately 1–2\% on target datasets.
\end{itemize}
\section{Related Work}
\label{sec:related}
Video Summarization can be broadly classified into: \textit{(i)} Video-to-Video summarization~\cite{bettadapura2016leveraging,he2023align,song2015tvsum}, where the task is to select important frames in the video to create a concise summarized video; and \textit{(ii)} Video-to-text summarization~\cite{han2023shot2story20k,hua2024v2xum}, with the objective of generating a text description that describes all the relevant information in a video in a concise manner.

\noindent\textbf{Video-to-Video Summarization} (hereinafter referred to as Video summarization, for clarity) approaches are further categorized into unsupervised learning~\cite{qiu2023liveseg,zhou2018deep,apostolidis2022summarizing,jung2019discriminative,de2011vsumm,narasimhan2022tl} and supervised learning~\cite{song2015tvsum,he2023align,apostolidis2021combining,fajtl2019summarizing,huynh2023cluster,park2020sumgraph,ji2019video,zhu2020dsnet,chaves2024videosage,otani2017video}.
Supervised methods use datasets such as TVSum~\cite{song2015tvsum} and SumMe~\cite{gygli2014creating}, which are smaller in size and lacking diversity.
In contrast, most unsupervised methods rely on clustering~\cite{narasimhan2022tl,de2011vsumm} techniques, where the frames are clustered based on either frame similarity or video-text alignment scores. 
IV-Sum\cite{narasimhan2022tl} trains on pseudo summaries generated from a large corpus of instructional videos through task relevance scores, and steps that are relevant to the task appear across many videos. 
Mr.HiSum \cite{sul2024mr}, on the other hand, uses the most replayed statistics from YouTube for training.
While these stats are synonymous to highlights, Sul \textit{et al.} show that it transfers well to video summarization tasks via a zero-shot setting.
Our proposed work builds on this approach and uses the most-replayed statistics for supervision, in addition to using the available video-level information and step-level instruction data during training. 

\noindent\textbf{Multi-Modal Summarization} aims at using multiple modalities to provide additional context to the model, leading to improved summarization. 
Existing works~\cite{bettadapura2016leveraging,li2018read,li2017multi,wu2022intentvizor} utilize complementary information from optical flow, audio, etc., to enhance summarization. 
In particular, language-guided frameworks have seen increasing interest in  recent literature~\cite{narasimhan2021clip,he2023align,narasimhan2022tl,li2017multi,plummer2017enhancing}.
To enable such video-text learning for summarization, VideoXum~\cite{lin2023videoxum, han2023shot2story20k, wu2024video} and UBiSS~\cite{mei2024ubiss} (building on ActivityNet and QVH datasets), created datasets for Video-Video and Video-Text summarization.
While these datasets provide text summaries and overviews, we focus on instructional videos from WikiHow that provides step-by-step instructions and also narrations with every video. 

The effectiveness of multimodal learning largely depends on how well video and text features are aligned.
Since the introduction of Transformers and CLIP~\cite{vaswani2023attentionneed,radford2021learning}, Transformer based Vision-Language models have been the go-to methods~\cite{videobert,yang2020bert,alamri2022end,beedu2025text,dong2024mamba,he2023align,zala2023hierarchical,guo2024vtg,shvetsova2022everything,plummer2017enhancing,haresamudram2025limitations}.
However, for video summarization using local step-by-step and global instructions, the model must align with the step-by-step instructions, and the global description of the video. 
HierVL~\cite{ashutosh2023hiervl}, learns a hierarchical parent-child model that simultaneously accounts for both long-and short-term data. 
Such hierarchical learning has been explored broadly in vision~\cite{yang2021hierarchical,feichtenhofer2019slowfast,zala2023hierarchical,li2020hero}, including video summarization in particular~\cite{zhao2018hsa,Yu2024HierarchicalMV}.
In this paper, we train a hierarchical network by utilizing clip-level subtitles and global instructions in a parent-child training strategy.
This approach enables the model to capture fine-grained details from subtitles while also aligning with the broader context provided by the video-level instructions.
\section{Methodology}
\label{sec:method}
\begin{figure*}
    \centering
    \includegraphics[width=0.7\textwidth]{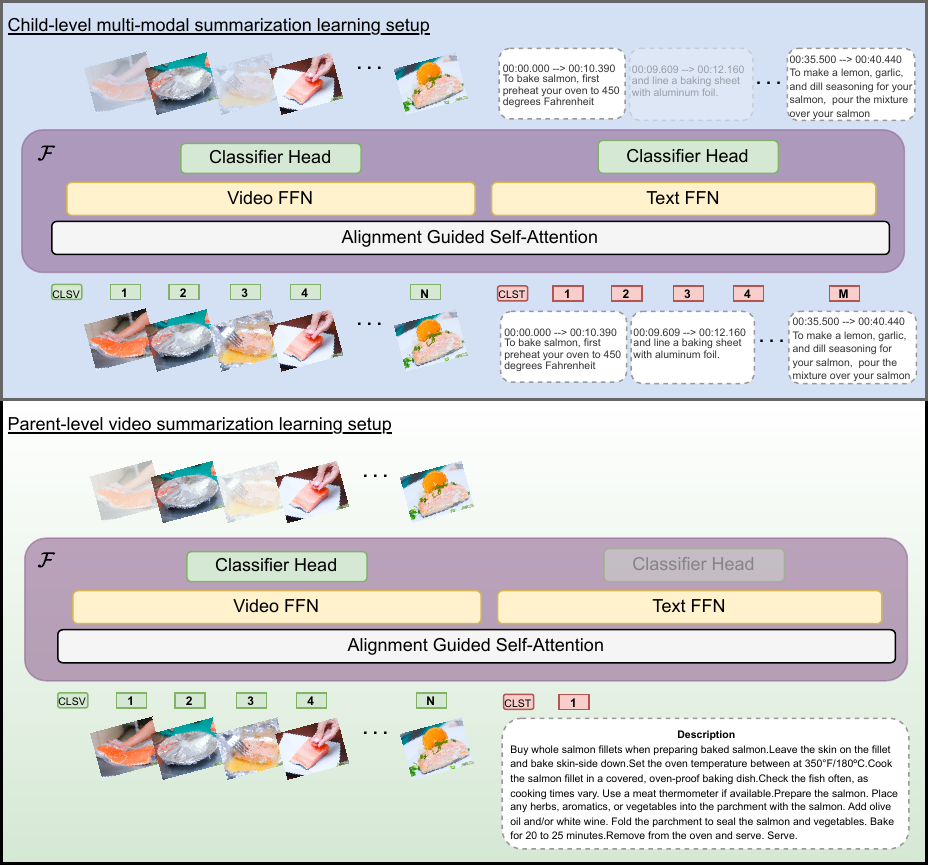}
    \caption{\textbf{Overview of HierSum:} in the sub-clip level (child level) training, with $\mathcal{N}$ frames and $\mathcal{M}$ subtitles as input, the model predicts the important frames and sentences as summaries. During the parent-level training, the subtitles are replaced with global descriptions and the model is trained only to predict important video frames. Note that the model $\mathcal{F}$ is common and is trained in both stages. When training with \textsl{most replayed} scores, the classifier head predicts the scores for each frame. }
    \label{fig:method}
\end{figure*}

In this paper, we propose HierSum -- a multi-modal architecture that takes video and text as inputs, as illustrated in \autoref{fig:method}, and generates video and text summaries.
It utilizes two levels of instructions, referred to as the parent and child instructions, emulating scenarios where adults typically need general directions, while children benefit from fine-grained step level instructions.
We leverage instructional videos, which typically contain sub-clip level information from narration/ASR, as well as global descriptions in the form of recipes/instructions.

Thus, the inputs for the model comprise $\mathcal{V}$ video frames, $\mathcal{T_{L}}$ subtitles, and  $\mathcal{T_{G}}$ descriptions.
From here on, for clarity, we use subtitles to denote the narration/ASR text input, and descriptions for any global instructions.

\paragraph{Task formulation:} 
Consider a video sample in a hierarchical dataset, 
\begin{math} 
\mathcal{D} = \{(\mathcal{V}_{i=1}^N, \mathcal{\{T_L\}}_{j=i}^{M}, \mathcal{T_G})\}, 
\end{math}
with $\mathcal{N}$ frames and $\mathcal{M}$ subtitles, and one global description. 
In line with prior works~\cite{he2023align,sul2024mr,zhu2020dsnet}, we use pre-extracted video features (e.g., CLIP Encoder \cite{radford2021learning}) and text features (e.g., SentenceTransformer \cite{reimers-2019-sentence-bert}) for each frame and subtitle and descriptions. 

Features from different modalities are projected onto a common embedding space using a linear layer.
We add a special token ``[CLS]" at the start of the feature sequence, and then add a learnable position embedding to each feature sequence to incorporate the order of the sequence. 
Each sentence that denotes the start and end of every subtitle is also embedded with timestep information, using a segment-based position embedding. 
The final input sequence consists of the video and text features, concatenated with positional embeddings. 
It is denoted as $\mathcal{X} \in \mathcal{R}^{(M+N)\times D}$ where $\mathcal{D}$ is the dimension of the features.

\paragraph{Hierarchical Setup:}
The overarching insight is that subtitle-level representations capture fine-grained actions relevant to a specific task in the video, whereas global instructions (in the form of description) contain the overall goal of the task.
For this, we develop an alternating training protocol where we train \textit{m} batches of subtitle-level visual and textual pairs, followed by one batch of a global text and visual pair.
Consider ``How to play UNO'' on WikiHow: the subtitle instructions consist of sentences such as \textit{``To play UNO, you'll need at least two players. You start by dealing 7 cards to each player ...''}. 
However, Wikihow instructions have the first step as \textit{``shuffle the cards and distribute 7 cards to each player''}. 
If one ignored the global instructions, the dealer might not shuffle the cards. 
We note that a caveat of using long descriptions during training is that capturing long-range sequences in video and text is typically challenging for transformer-based architectures~\cite{devlin2019bert}.
Yet, Sentence transformer models~\cite{reimers-2019-sentence-bert} generally perform well for long sentences. 
We specifically use the Sentence-BERT (S-BERT) model for generating text representations.

\paragraph{Feature Alignment:}
One core requirement is to align the video features with the relevant subtitle so that the subtitles can be used to learn the features effectively. 
A2Summ~\cite{he2023align} proposes an attention-guided self-attention mechanism where masking is introduced over subtitles. 
Specifically, an attention mask $\mathcal{A} \in \mathcal{R}^{(N+M) \times (N+M)}$ initialized with $0$ is defined to indicate the alignment of the time step.   
We fill the mask with ones corresponding to the same segment to ensure cross-modal attention between the video and text inputs.
The attention mask is then applied to the attention matrix~\cite{he2023align}.

\paragraph{\textsl{Most Replayed} Statistics:} The \textsl{most replayed} statistic on YouTube is a publicly visible feature that shows the frequency of ``rewatched'' views~\cite{Google}.
These statistics tend to be more accurate when views exceed $50k$. 
We posit that there is a correlation between the highlights and the video summarization.  
As we require instructional videos, we crawl WikiHow and EHow videos on YouTube and gather all videos with $>50k$ views that have a corresponding instructional website. 
We use this dataset to train a model that predicts the relevancy score and show that it is a highly effective pre-training protocol for video summarization.

\paragraph{Loss Function:} The final loss  comprises: \textit{(i)} Classification loss; and \textit{(ii)} Contrastive loss. \\
\noindent\emph{(i) Classification Loss}. We apply focal loss~\cite{lin2017focal} to classify the importance score.

\small{
\begin{equation}
    L_\text{cls}=
    \frac{1}{N}\sum_{i=1}^{N}\left\{
    \begin{matrix}
         -\alpha(1-p_i)^\gamma \log(p_i), & \mbox{if} & y_i=1 \\
         -(1-\alpha)p_i^\gamma \log(1-p_i), & \mbox{if} & y_i=0
    \end{matrix}\right. 
\end{equation}
}
\normalsize {
\begin{equation}
L_\text{cls} = L_{\text{cls}_\text{video}} + L_{\text{cls}_\text{text}}
\end{equation}
}

where $p_i$ is the predicted score for each frame and sentence, and $y_i$ is the ground-truth label where $y_i=1$ indicating \textit{i$^{th}$} frame as the keyframe/key-sentence. 
During the global step, only $L_{\text{cls}_\text{video}}$ is used to train. 

While the focal loss is applied for the importance labels, as we also utilize the most-replayed statistics during training, we employ a mean-squared error loss on the relevancy scores. 
All frames with relevancy scores $\ge$ 0.15 are considered ``important" and labeled as 1.
$$\mathcal{L}_{mse} = \frac{1}{N}\sum_{i=1}^{N} ||s_i - p_i||^2$$
where $s_i$ is the ground-truth relevancy score.

\noindent\emph{(ii) Contrastive Losses}: Following A2Summ~\cite{he2023align} and driven by the success of contrastive learning~\cite{radford2021learning,beedu2025text,choi2023multimodal,haresamudram2025limitations}, we design an auxiliary contrastive loss, $\mathcal{L}_\text{inter}$, by maximizing the cosine similarity of the video embedding [CLSV] and text embedding [CLST] from $B$ positive pairs from the batch, and $B^2 - B$ negative samples.
While prior methods employ inter-modality contrastive loss, A2Summ~\cite{he2023align} proposes an intra-modality loss for video summarization task that is able to distinguish keyframes and sentences from the background frames and less-related sentences.

To achieve this, the wrongly classified frame/sentence with high prediction scores are selected as hard-negative samples, after excluding the samples that are close to positive keyframe samples to avoid confusing the model, and the pre-defined ground-truth samples are considered as positive samples. 
For more information on intra-sample contrastive loss, we refer readers to~\cite{he2023align}.
The loss is defined as:
\begin{equation}
\begin{aligned}
    \mathcal{L}_\text{intra} = \mathbb{E}_{z \sim \mathcal{I}_{PF}, z+\sim \mathcal{I}_{PS}, z-\sim \mathcal{I}_{HNF}}l(z,z^+,z^-) \\
    + \mathbb{E}_{z \sim \mathcal{I}_{PS}, z+\sim \mathcal{I}_{PF}, z-\sim \mathcal{I}_{HNS}}l(z,z^+,z^-)
\end{aligned}
\end{equation}
Where $\mathcal{I}_{PF}, \mathcal{I}_{PS}, \mathcal{I}_{HNF}, \mathcal{I}_{HNS}$ are positive frames, positive sentences, hard-negative frames and hard-negative sentences.

The final loss is defined as:
\begin{equation}
    \mathcal{L} = \mathcal{L}_\text{cls} + \alpha \mathcal{L}_\text{mse} + \beta \mathcal{L}_\text{inter} + \lambda \mathcal{L}_\text{intra}
\end{equation}
where $\alpha$, $\beta$ and $\gamma$ are hyper-parameters controlling the trade off between the losses.
\section{Experiments}
\label{sec:exp}
\subsection{Datasets}
We evaluate our approach on four datasets, namely: TVSum~\cite{song2015tvsum}, BLiSS~\cite{he2023align}, Mr.HiSum~\cite{sul2024mr}, and WikiHow~\cite{narasimhan2022tl}. \\

\noindent \textbf{TVSum} contains 50 videos spanning 10 categories (how-to videos, news, documentaries, etc.) with five videos per category.
Video summaries are constructed for each user annotation using the 0/1 knapsack algorithm.
\textbf{BLiSS} is a multimodal summarization dataset consisting of 628 live streamed videos collected from Behance, where text and video are temporally aligned. 
Each video is segmented into 5-minute clips and is annotated for key-sentences. 
\textbf{Mr.HiSum} is a video highlight and summarization dataset containing $\sim$ 32k videos from the YouTube-8M dataset.
Each video is typically between 120 and 300 seconds in length, with an average of 202 seconds. 
This dataset leverages the \textsl{most replayed} statistics as importance scores, which are then aggregated to shot-level using KTS~\cite{potapov2014category} boundary information. 
\textbf{WikiHow Summaries}~\cite{narasimhan2022tl} is an instructional video summarization dataset consisting of 2106 videos where each video describes a unique task. 
These videos are obtained from the wikiHow website along with step-by-step instructions.
We use this dataset as test bed, similar to \cite{narasimhan2022tl}.

\begin{table}[!t]
    \centering
    \resizebox{0.9\columnwidth}{!}{
    \begin{tabular}{C{.4\linewidth} C{.15\linewidth} C{0.2\linewidth} C{0.2\linewidth} C{0.1\linewidth}}
    \toprule
    Dataset           &  {\# Videos} & {Total Time (hr)} & {Avg. Time (min)} & Desc. \\ 
    \midrule 
    Mr. HiSum         & 31892 & 1788 & 3.36  & S \\ 
    TL;DW (Pseudo)     & 12160 & 628.5 & 3.1  & S \\ 
    LfVS-P            & 250K  & 55K & 13.30  & S \\ 
    TL;DW (WikiHow)   & 2106  & 42.9 & 1.23  & G \\ 
    \midrule
    Ours          & 2413  & 74.8 & 1.86      & S,G\\ 
    \bottomrule
    \end{tabular}
    }
    \captionsetup{type=table}
    \captionof{table}{Dataset comparison. S - Subtitles, G - Global descriptions.}
    \label{tab:dataset}
\end{table}

\subsection{Instructional Video Data Collection}
While existing datasets for video summarization are extensive and diverse, they lack two-level text that integrates both local sub-clip level information and global instructions, particularly for instructional videos. 
To address this limitation and inspired by~\cite{narasimhan2022tl}, we curate a dataset containing instruction videos from the WikiHow and EHow \footnote{\url{https://www.wikihow.com}, \url{https://www.ehow.com/}} platforms.
Specifically, we scrape their respective YouTube channels and collect videos that are both linked to instructional articles and have over 50k views.
Following MrHiSum~\cite{sul2024mr}, we prioritize videos that have most-replayed statistics.
Each selected instructional video is either accompanied by a detailed description in the YouTube footnotes or contains stepwise instructions in the corresponding article. 
Additionally, we extract the subtitles using yt-dlp. 
For videos lacking cleanly generated subtitles, we employ Whisper~\cite{radford2023robust} to extract text directly from the audio.
In cases where a video lacks a corresponding article, we utilize its YouTube description as the global instruction.

\begin{figure}[!t]
\centering
    \includegraphics[width=1\linewidth]{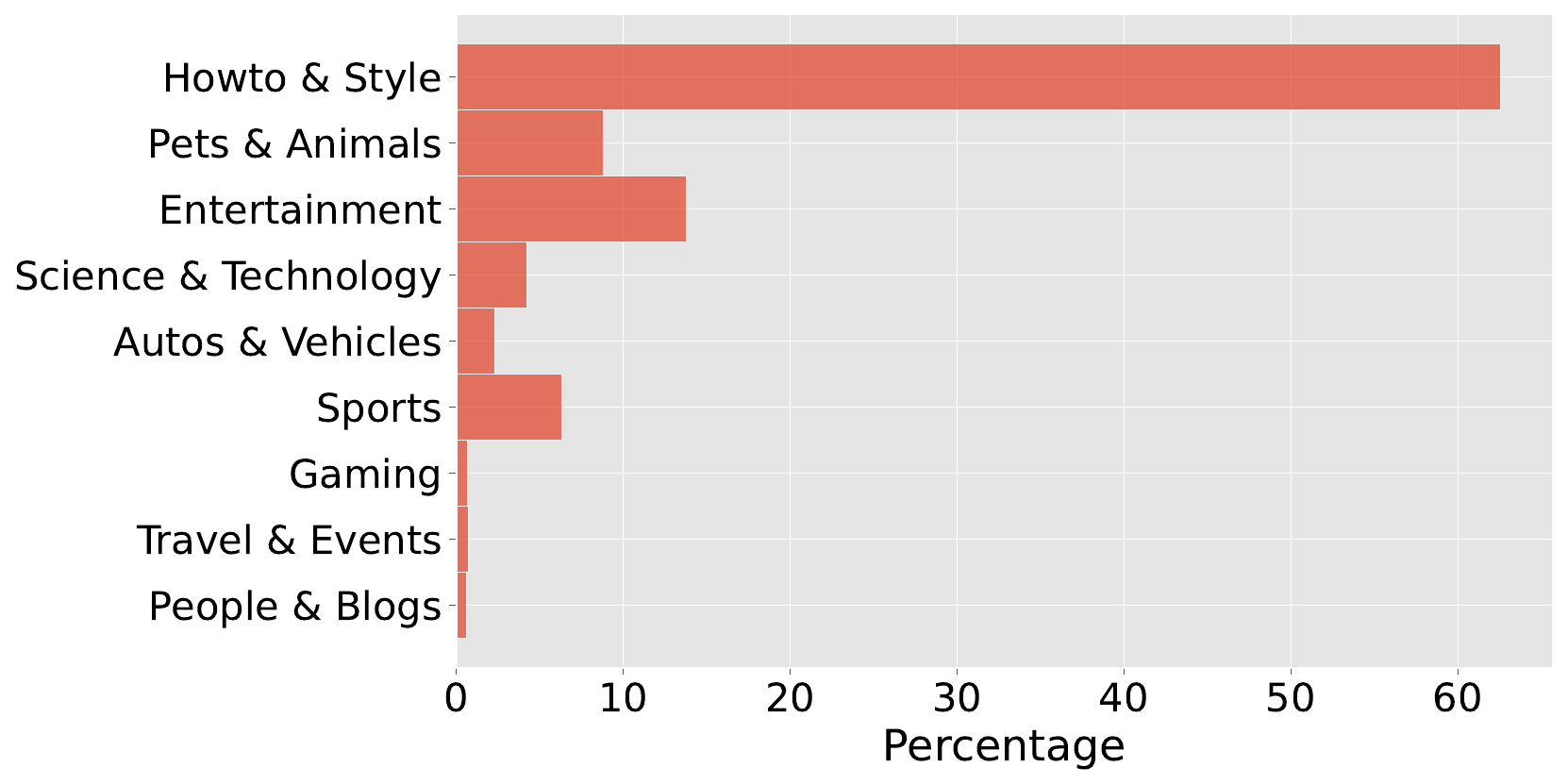}
    \captionof{figure}{Distribution of categories in the dataset.}
    \label{fig:tasks}
\end{figure}

We compare our dataset against existing datasets in \autoref{tab:dataset}. 
While our dataset has 9 main categories (\autoref{fig:tasks}), ``HowTo \& Style" spans diverse sub-tasks such as gardening, crafts, sawing, and decor.
Although TL;DW~\cite{narasimhan2022tl} also leverages the WikiHow dataset for evaluation, our pre-training dataset is unique in providing multi-granularity descriptions.
Unlike previous datasets, ours includes both sub-clip-level transcriptions and human-written global instructions, which we argue are \textbf{both essential} for effective video summarization.
To the best of our knowledge, this is the first dataset to provide text data at both levels.
The dual-level annotations in our dataset serve as a key differentiator, providing richer supervision during training and facilitating more effective modeling of both local and global video contexts, leading to performance gains that go beyond simply scaling the pre-training dataset.

\subsection{Evaluation}
For TVSum and WikiHow, following previous work~\cite{he2023align,sul2024mr,chaves2024videosage,narasimhan2021clip}, we evaluate summarization using F1-score, and rank order statistics (Kendall's $\tau$ and Spearman's $\rho$)~\cite{otani2017video}.
The generated text summary for BLISS is evaluated using ROUGE \cite{lin2004rouge}, specifically the R-1, R-2, and R-L scores, whereas the cosine image similarity is used for video summaries. 
In the case of the Mr.HiSum dataset, similar to \cite{sul2024mr}, we report F1-scores for summarization and the Mean Average precision (MAP) scores for highlight detection.
\
\subsection{Implementation details}
For TVSum, we follow prior works~\cite{he2023align,sul2024mr,chaves2024videosage} and use pre-extracted GoogleNet features as video input. 
In the case of BLISS and text data for TVSum, we follow A2Summ~\cite{he2023align} and use the pre-extracted features as inputs.
BLISS does not have global instructions, and we utilize the key-sentences as the text input during the parent learning step. 
For all text data (save the TVSum dataset), we use the SentenceTransformer's BERT encoder~\cite{reimers-2019-sentence-bert} to extract features.
Inception-v3~\cite{szegedy2016rethinking} features are extracted for all videos from the Mr.HiSum dataset.
In contrast, for the Wikihow (testbed) dataset as well as our instructional dataset, we use the pre-trained CLIP~\cite{radford2021learning} Encoder to extract features for each frame, sampled at 1 FPS.
During evaluation, we use only the clip-level subtitles as the text input.


\subsection{Results}
\textbf{TVSum Dataset: }
We compare HierSum to state-of-the-art methods on TVSum~\cite{song2015tvsum} dataset in \autoref{tab:tvsum}.
It achieves the best performance for rank coefficients, while slightly under-performing on the F1-score.
Yet, HierSum is the better option, as Otani \etal~\cite{otani2017video} argue that rank coefficients are a better measure for evaluating how close the generated summaries are to human annotated scores. 

HierSum shows improvements of 2\% for the Kendall's $\tau$ metric and by 5\% for Spearman's $\rho$ metric over baselines, demonstrating the capability to effectively summarize videos.
We note that VideoSage~\cite{chaves2024videosage} uses a different evaluation protocol, whereas V2Xum-LLaMa~\cite{hua2024v2xum} trains a Llama model and has an architecture substantially different from ours, thereby rendering  comparison challenging.
However, for completeness, we also report these scores.

\begin{table}[!t]
\def\arraystretch{1.2}
\centering
\setlength{\tabcolsep}{1mm}
\resizebox{0.8\columnwidth}{!}{
\begin{tabular}{C{.4\columnwidth} C{.2\columnwidth} C{0.2\columnwidth} C{.2\columnwidth} }
\toprule
Method      & F1 score  & $\tau$ & $\rho$ \\
\midrule
\textcolor[HTML]{656565}{V2Xum-LLaMa} &  \textcolor[HTML]{656565}{-}   & \textcolor[HTML]{656565}{0.222}  & \textcolor[HTML]{656565}{0.293}  \\ 
\textcolor[HTML]{656565}{VideoSage}   &  \textcolor[HTML]{656565}{-}   & \textcolor[HTML]{656565}{0.3}    & \textcolor[HTML]{656565}{0.42}   \\ 
\midrule
CLIP-It \cite{narasimhan2021clip}     & 64.2 & 0.108  & 0.147  \\
TL;DW \cite{narasimhan2022tl}       &  -   & 0.143  & 0.167  \\
A2Summ \cite{he2023align}      & 63.4 & 0.150  & 0.178  \\
LfVS\cite{argaw2024scaling} & \textbf{72.38} & 0.169 & 0.203 \\
CSTA (GoogleNet)\cite{son2024csta} & - & 0.129 & 0.163 \\
\midrule
Ours        & {62.5} & \textbf{0.172}  & \textbf{0.225}  \\
\bottomrule
\end{tabular}
}
\caption{\textbf{TVSum}: Comparison with state-of-the-art methods for F1-score, Kendall's $\tau$ and Spearman's $\rho$ metrics. We include the results of using GoogleNet's features for fair comparison. 
While LfVS performs the best for F1 metric, our method outperforms significantly for rank coefficients.
}
\label{tab:tvsum}
\end{table}

\begin{table}[!t]
\def\arraystretch{1.2}
\centering
\setlength{\tabcolsep}{1mm}
\resizebox{0.8\columnwidth}{!}{
\begin{tabular}{C{.3\columnwidth} C{.15\columnwidth} C{0.15\columnwidth} C{.15\columnwidth}  C{.15\columnwidth} }
\toprule
Method   & R-1   & R-2   & R-L       & Cos(\%) \\
\midrule
DSNet-AF \cite{zhu2020dsnet} & -     & -     & -         & 62.7    \\
CLIP-It \cite{narasimhan2021clip}  & -     & -     & -         & 63.58   \\
BART \cite{lewis2019bart}     & 49.11 & 38.59 & 48.08     & -       \\
A2Summ$^*$ \cite{he2023align}  & 52.47  & 41.7  & 51.44     & 63.65   \\
\midrule
Ours     & 52.3  & 41.5 & 51.2      & \textbf{64.8 }  \\
\bottomrule
\end{tabular}
}
\caption{\textbf{BLiSS}: Performance comparison for R-1, R-2, R-L and cosine similarity metrics with state-of-the art methods. Our method achieves the best performance on BLiSS dataset.}
\label{tab:bliss}
\end{table}

\noindent \textbf{BLiSS Dataset: }
We validate HierSum on the livestream dataset -- BLiSS, in \autoref{tab:bliss}, by comparing both text and video summarization performances.
It achieves the best performance for video cosine similarity, indicating improvements for video summarization. 
As BLISS does not have ``global'' descriptions available, we instead used key-phrases in lieu of these descriptions. 
We believe that by not training for text summarization at every global step negatively impacts overall performance for text summarization, where HierSum's performance is slightly worse than A2Summ.
We also note that A2Summ's performance was obtained by retraining its model using the official code to eliminate any version/setup dependent inaccuracies. 

\noindent \textbf{Mr.HiSum Dataset: }
We tabulate the performance of our model performance on the Mr.HiSum dataset in \autoref{tab:mrhisum}. 
By incorporating multiple modalities and training on our instructional data, we show  improvements $\ge 1\%$ on the F1-score, $\sim2\%$ on MAP when the top 50\% shots are used and more than 5\% when only top 15\% shots are used. 
While our model is specifically trained to predict the \textsl{most-replayed} scores (i.e., the  highlight moments), we see that it is effective at summarization as well, indicating the effectiveness of using the most-replayed scores (if  available).

\begin{table}[!t]
\def\arraystretch{1.2}
\centering
\setlength{\tabcolsep}{1mm}
\resizebox{0.8\columnwidth}{!}{
\begin{tabular}{C{.3\columnwidth} C{.15\columnwidth} C{0.275\columnwidth} C{.275\columnwidth} }
\toprule
Method      & F1  & MAP\footnotesize${\rho = 50\%}$ & MAP\footnotesize${{\rho = 15\%}}$ \\
\midrule
PGL-SUM \cite{apostolidis2021combining}   & 56.89 & 62.02  & 27.71  \\
VasNet \cite{fajtl2019summarizing}     & 55.26 & 58.69  & 25.28  \\
SL-module \cite{xu2021cross}      &  55.31   & 58.63  & 24.95  \\

\midrule
Ours        & \textbf{58.16} & \textbf{63.8}  & \textbf{32.6}  \\
\bottomrule
\end{tabular}
}
\caption{\textbf{Mr.HiSum}: We compare F1 score, MAP at $\rho$ = 50\% and  $\rho$ =15\% of our method to all the baselines. Our method outperforms all other baseline methods across all metrics.}
\label{tab:mrhisum}
\end{table}

\noindent \textbf{WikiHow Dataset: }
In \autoref{tab:wikihow}, we evaluate our method on WikiHow Summaries~\cite{narasimhan2022tl}. 
Among baselines, Frame Cross-Modal Similarity has the closest model configuration to ours. 
All other methods (\textcolor[HTML]{656565}{grayed out}) use segments of 32 frames sampled at 8 FPS, or convert frame-level scores to shot-level scores using the 0/1 knapsack algorithm.
For fair comparison, we also train our model on Pseudo Summaries provided by \cite{narasimhan2022tl} using local and global instructions, and compare against training on our dataset. 
We generate summaries by selecting the top 55\% of the highest scoring frames, similar to \cite{narasimhan2022tl}. 
Training our model on the \textsl{most replayed} statistics, as opposed to the pseudo summaries, results in $\sim$2 \% improvement for F1-score, and a substantial 3\% improvement on the rank coefficients. 

\begin{table}[!t]
\def\arraystretch{1.2}
\centering
\setlength{\tabcolsep}{1mm}
\resizebox{0.9\columnwidth}{!}{
\begin{tabular}{C{.6\columnwidth} C{.1\columnwidth} C{0.15\columnwidth} C{.15\columnwidth} }
\toprule
Training Dataset & F1  &$\tau$ & $\rho$  \\
\midrule
\textcolor[HTML]{656565}{Frame Cross-Modal Similarity} & \textcolor[HTML]{656565}{53.1} & \textcolor[HTML]{656565}{0.022} & \textcolor[HTML]{656565}{0.051} \\
\textcolor[HTML]{656565}{Step Cross-Modal Similarity} & \textcolor[HTML]{656565}{58.3} & \textcolor[HTML]{656565}{0.037} & \textcolor[HTML]{656565}{0.061} \\
\textcolor[HTML]{656565}{CLIP-It with ASR} & \textcolor[HTML]{656565}{62.5} & \textcolor[HTML]{656565}{0.093} & \textcolor[HTML]{656565}{0.191} \\
\textcolor[HTML]{656565}{IV-Sum~\cite{narasimhan2022tl}} & \textcolor[HTML]{656565}{67.3} & \textcolor[HTML]{656565}{0.101} & \textcolor[HTML]{656565}{0.212} \\
\midrule
Ours (trained on Pseudo)   & 46 & 0.05 &0.07  \\
Ours (trained on WikiHow)   & 48.17  &0.08 & 0.1  \\        

\bottomrule                           
\end{tabular}
}
\caption{Model performance on the WikiHow test dataset}
\label{tab:wikihow}
\end{table}

\section{Ablation Studies}
\subsubsection*{Cross-Dataset Verification}
To further investigate the contributions of the parent-child learning protocol using local subtitles and global descriptions, and to demonstrate that the \textsl{most replayed} statistics in combination with text modalities are an effective learning protocol, we perform cross-dataset validation on BLiSS and Mr.HiSum datasets in \autoref{tab:cross_dataset_bliss} and \autoref{tab:cross_dataset_hisum}.
It is also worth noting that during cross-data verification, the features are extracted using different encoders. 

\noindent\textbf{BLiSS:} 
The baseline model for the BLiSS dataset was trained with a hidden size of 128, whereas our model was trained with a larger hidden size of 512, which we denote as BLiSS(L).
We observe that for optimal performance on this dataset, the smaller model configuration is preferable. 
However, upon fine-tuning our larger model pre-trained using \textsl{most replayed} statistics along with global and local instructions, we see an improvement of 1\% in performance. 
We note the zero shot performance of the model (denoted as ZS) and notice that it it performs comparably, within a margin of 5\% with fully trained model.

\noindent\textbf{Mr HiSum:} Similar observations can be made for Mr.HiSum as well, which is also trained using the \textsl{most replayed} statistics in \autoref{tab:cross_dataset_hisum}.
While it could be expected that zero shot performance would be closer to end-to-end trained model, given that both these datasets are trained to predict the most replayed scores, we theorize that a domain gap exists in the extracted input features.
Specifically, Mr.HiSum dataset uses features extracted from the Inception model, whereas our model extracts features from the CLIP Encoder. 
However, this performance gap is reduced when the model is fine-tuned on the target dataset.

\begin{table}[!t]
\centering
\def\arraystretch{1.2}
\centering
\setlength{\tabcolsep}{1mm}
\resizebox{0.8\columnwidth}{!}{
  \begin{tabular}{C{.4\columnwidth}C{.4\columnwidth}C{.2\columnwidth} }
    \toprule
\multicolumn{2}{c}{Dataset} & Cos(\%)  \\ \cline{1-2}
Training     & Test         &       \\       \midrule
BLiSS        & BLiSS        & 63.7  \\
BLiSS(L)     & BLiSS(L)  & 62.0  \\
HierSum      & BLiSS(ZS)    & 59.2  \\
HierSum      & BLiSS(L)     & \textbf{64.7}  \\
    \bottomrule
  \end{tabular}
  }
  \caption{Video Summarization performance on Cross-dataset transfer on BLiSS dataset}
\label{tab:cross_dataset_bliss}
\end{table}

\begin{table}[!t]
\centering
\def\arraystretch{1.2}
\centering
\setlength{\tabcolsep}{1mm}
\resizebox{0.9\columnwidth}{!}{
  \begin{tabular}{C{.2\columnwidth}C{.2\columnwidth}C{.1\columnwidth}C{.25\columnwidth}C{.25\columnwidth} }
    \toprule
\multicolumn{2}{c}{Dataset} & F1  & MAP\footnotesize${\rho = 50\%}$ & MAP\footnotesize${\rho = 15\%}$ \\ \cline{1-2}
Training     & Test         &       \\       \midrule
HiSum        & HiSum        & 57.6 & 62.4  & 32.6  \\
HierSum      & HiSum(ZS)   & 44.7 & 57.1  & 23.0   \\
HierSum      & HiSum        & \textbf{57.7} & \textbf{62.7}  & \textbf{32.9 }  \\
    \bottomrule
  \end{tabular}
  }
\caption{Video Summarization performance on Cross-dataset transfer on Mr.HiSum dataset}
\label{tab:cross_dataset_hisum}
\end{table}

\begin{figure}[!t]
    \centering
    \includegraphics[width=0.75\linewidth]{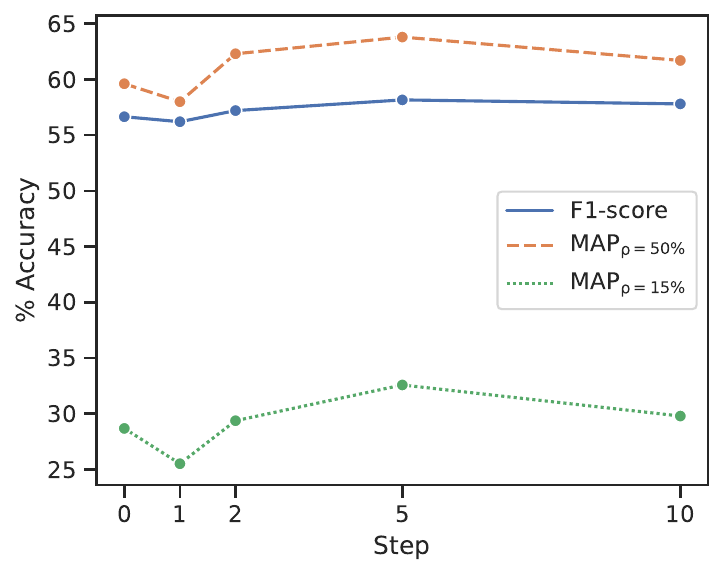}
    \caption{Performance comparison for different global and local training steps on Mr.HiSum. Global step=5 indicates that for every five samples of training with local subtitles, a sample with global description is trained. ASR only indicates that the model was trained only using the local subtitles.}
    \label{fig:global_step_ablation}
\end{figure}

\subsubsection*{Training protocol}
In \autoref{fig:global_step_ablation}, we investigate how often our model should train with global instructions in the Mr.HiSum dataset.
$Step = 0 $ indicates that the model was trained exclusively using subtitles, without incorporating any global instructions. 
$Step = 1$ indicates that only the global instructions were used during training.
While the model benefits from subtitles/ASR inputs, its performance improves when global instructions are introduced once every 5 local steps, compared to every 2 local steps. 
We hypothesize that although global instructions enhance the model's learning, introducing it at a higher frequency overwhelms the model, making it challenging to align each video frame with the overarching instructions. 
We further notice that increasing this step interval to 10 results in a decline in performance. 
These results suggest that the optimal parent-child learning step may vary depending on the dataset and the instructions.
However, integrating instructions at both local and global levels remains crucial for enhancing model performance.

\begin{figure*}[t!]
    \centering
    \begin{subfigure}[t]{1\textwidth}
        \centering
        \includegraphics[width=1\textwidth]{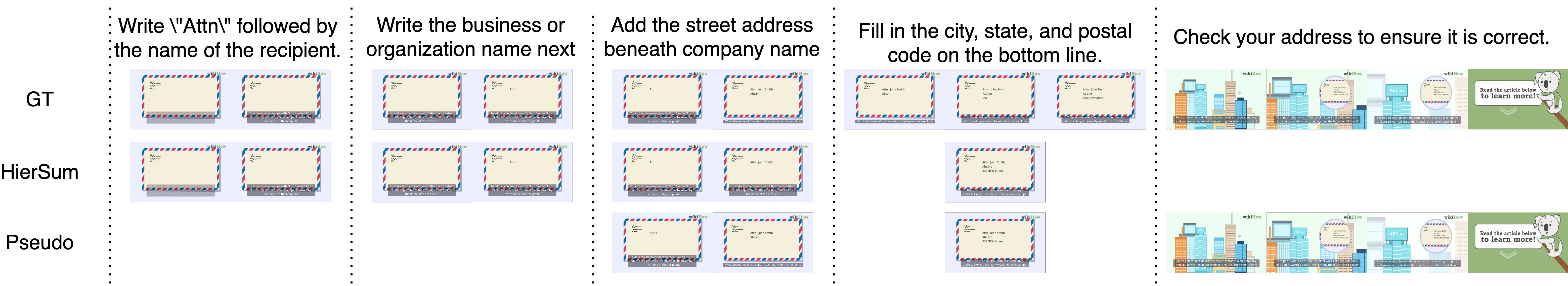}
        \caption{Address-Envelopes-With-Attn}
        \label{fig:example1}
        \hspace{1em} 
    \end{subfigure}%
    \\ 
    \begin{subfigure}[t]{1\textwidth}
        \centering
        \includegraphics[width=1\textwidth]{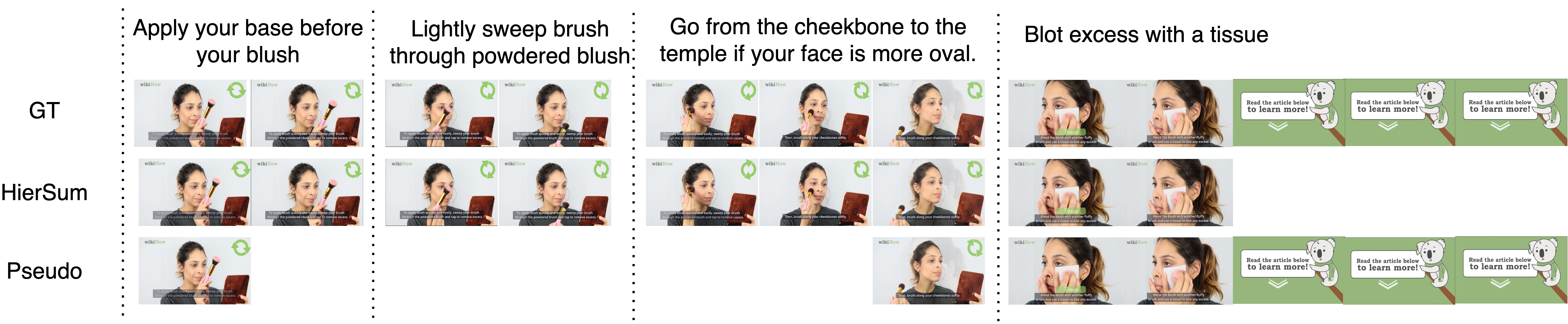}
        \caption{Apply-Blush}
        \label{fig:example2}
        \hspace{1em}
    \end{subfigure}
    \\
    \begin{subfigure}[t]{1\textwidth}
        \centering
        \includegraphics[width=1\textwidth]{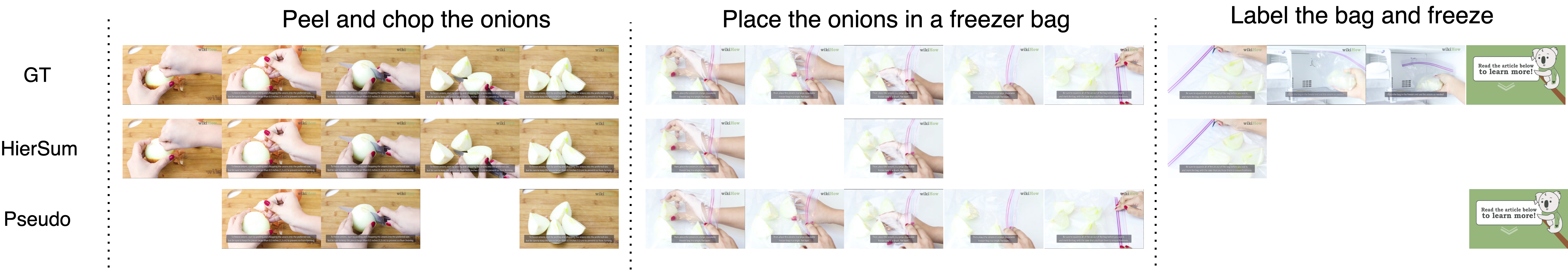}
        \caption{Freeze-Onion}
        \label{fig:example3}
    \end{subfigure}
    \caption{\textbf{Qualitative Results.} We show summaries from our method HierSum trained on Pseduo Summaries~\cite{narasimhan2022tl} and our dataset. HierSum, when trained on our dataset assigns higher scores to all the frames relevant, and lower scores that aren't crucial to the task, e.g WikiHow logo. We note a failure case in (c).}
    \label{fig:qualitative}
\end{figure*}

\section{Qualitative results}
We present qualitative results in \autoref{fig:qualitative}, comprising the  summaries generated using ground truth information, HierSum trained on our dataset, and HierSum trained on pseudo summaries (for ease, we will refer to this as $HierSum\_Pseudo$ in this section). 
In \autoref{fig:example1}, $HierSum\_Pseudo$ misses the first two instructions, and retains all frames from the tail end of the video where the frames are not very relevant to the task. 
Similarly for \autoref{fig:example2}, $HierSum\_Pseudo$ misses some crucial steps such as ``\textit{Lightly sweeping your brush through the blush}" and ``\textit{Go from cheekbone to temple}".
We also observe that the ground truth summaries contain a few redundant frames, and the wikiHow logo is present in all of them. 
It is worth noting that while HierSum generated better summaries, the F1-score for this video was higher for $HierSum\_Pseudo$ (81\% vs 27\%). 
That is a failure case of using the ground truth summaries.
We also note a failure case of HierSum, where the F1 score for ``\textit{Freeze-Onion}" was 90\% as opposed to 55\% for $HierSum\_Pseudo$, but HierSum missed an important step of placing the onions in the freezer.
\section{Conclusion}
\label{sec:conclusion}
We introduce a novel approach for generating visual summaries by using a hierarchical framework that effectively integrates fine-grained local cues from subtitles along with global video-level instructions. 
Our method leverages the ``most replayed” statistics as an additional supervisory signal, enabling the model to accurately identify and prioritize the critical segments of instructional videos. 
By using this statistics, in addition to the two-tier text information, our model is able to effectively learn to summarize videos.
Extensive evaluations on benchmark datasets, including TVSum, BLiSS, Mr.HiSum, and WikiHow, demonstrate that HierSum not only outperforms state-of-the-art approaches in key metrics such as F1-score and rank correlation, but also produces summaries that are more aligned with human judgment.
Our ablation studies confirm that the two-stage (parent-child) learning strategy is crucial for effectively capturing both local and global information, and cross-dataset validations underscore the robustness and adaptability of our approach across diverse video domains. 

{
    \small
    \bibliographystyle{ieeenat_fullname}
    \bibliography{main}
}


\end{document}